# Refine Medical Diagnosis Using Generation Augmented Retrieval and Clinical Practice Guidelines

Wenhao Li, Hongkuan Zhang, Hongwei Zhang, Zhengxu Li, Zengjie Dong, Yafan Chen, Niranjan Bidargaddi, Hong Liu*

***Abstract*—Current medical language models, adapted from large language models (LLMs), typically predict ICD code-based diagnosis from electronic health records (EHRs) because these labels are readily available. However, ICD codes do not capture the nuanced, context-rich reasoning clinicians use for diagnosis. Clinicians synthesize diverse patient data and reference clinical practice guidelines (CPGs) to make evidence-based decisions. This misalignment limits the clinical utility of existing models. We introduce GARMLE-G, a Generation-Augmented Retrieval framework that grounds medical language model outputs in authoritative CPGs. Unlike conventional Retrieval-Augmented Generation based approaches, GARMLE-G enables hallucination-free outputs by directly retrieving authoritative guideline content without relying on model-generated text. It (1) integrates LLM predictions with EHR data to create semantically rich queries, (2) retrieves relevant CPG knowledge snippets via embedding similarity, and (3) fuses guideline content with model output to generate clinically aligned recommendations. A prototype system for hypertension diagnosis was developed and evaluated on multiple metrics, demonstrating superior retrieval precision, semantic relevance, and clinical guideline adherence compared to RAG-based baselines, while maintaining a lightweight architecture suitable for localized healthcare deployment. This work provides a scalable, low-cost, and hallucination-free method for grounding medical language models in evidence-based clinical practice, with strong potential for broader clinical deployment.**

## I. INTRODUCTION

THE rapid development of artificial intelligence technologies, particularly large language models (LLMs), is leading human society into an unprecedented technological revolution, rapidly reshaping the working patterns and developmental landscapes of various traditional industries. In the healthcare sector, language models and related tools, such as ChatGPT and ClinicalBERT, have been increasingly applied across multiple scenarios, including disease prediction, clinical decision support, patient interaction, drug discovery, and personalized medicine, significantly driving innovation and transformation in medical technology [1, 2].

As a fundamental task in healthcare, disease diagnosis refers to the process by which health professionals identify the most likely disease or disorder causing a patient's symptoms [3]. An accurate diagnosis depends on factors such as the timing and progression of symptoms, the patient's medical history, risk factors for specific diseases, and any recent exposure to potential health risks. In recent years, with advancements in electronic health records (EHR), knowledge essential for making diagnoses—in the form of clinical notes, pathology reports, imaging results, and medical prescriptions, can now be accessed easily, while historical diagnoses are stored in the form of a series of ICD codes [4]. Based on historical EHR datasets, medical language models are trained, using which new EHR records can be quickly mapped to a series of ICD code-based diseases or disorders, aiming to improve the quality and efficiency of medical processes [1, 5, 6].

Despite the promise of medical language models, their integration into clinical practice remains limited by a fundamental disconnect: most models are optimized to generate ICD code-based outputs, while real-world clinical decision-making is a dynamic, iterative process grounded in the synthesis of diverse patient data and the application of clinical practice guidelines (CPGs) [7, 8]. Clinical workflows require not only the recognition of disease patterns but also the

The research reported in this paper is financially supported by the National Natural Science Foundation of China (62276156), the project of Shandong Provincial Natural Science Foundation (ZR2024LZH005), the Taishan Scholar Program of Shandong Province of China (No.tsqnz20240809), and the Excellent Youth Foundation of Shandong Natural Science Foundation (2024HWYQ-055). Corresponding author: Hong Liu.

Wenhao Li is with Shandong Normal University, Jinan, China, 250358 (email: lwh@sdnu.edu.cn)

Hongzang Li is with Shandong Normal University, Jinan, China, 250358 (email: 2024217028@stu.sdnu.edu.cn)

Hongwei Zhang is with Shandong Maternal and Child Health Hospital, Jinan, China, 250014 (email: zhanghw1999@163.com)

Zhengxu Li is with Shandong Normal University, Jinan, China, 250358 (email: 2024217058@stu.sdnu.edu.cn)

Zengjie Dong is with Shandong Normal University, Jinan, China, 250358 (email: 2024317115@stu.sdnu.edu.cn)

Yafan Chen is with Shandong Normal University, Jinan, China, 250358 (email: 2024217054@stu.sdnu.edu.cn)

Niranjan Bidargaddi is with Flinders University, Adelaide, Australia, 5042 (email: niranjan.bidargaddi@flinders.edu.au)

Hong Liu is with Shandong Normal University, Jinan, China, 250358 (email: hongliu@sdnu.edu.cn)

contextualization of findings within evolving evidence and patient-specific factors. CPGs serve as the backbone of these workflows, guiding nuanced, patient-centered decisions that adapt as new information emerges. When AI tools produce static, code-based predictions divorced from this process, they risk marginalizing the expertise and adaptability that define effective clinical care. This misalignment reduces the practical value of AI-driven solutions, impedes their adoption by clinical teams, and ultimately limits their impact on patient outcomes. Bridging this gap requires approaches that embed the reasoning and evidence structures of clinical workflows directly into model outputs, ensuring that AI augments—rather than bypasses—the complexity of real-world medical decision-making.

The most intuitive and cost-effective approach to eliminate this misalignment is to extend existing models to support CPGs. This involves integrating medical language models with these guidelines across various disease domains, thereby expanding model output from ICD code-based diagnoses to more comprehensive, guideline-driven diagnostic and treatment knowledge. This approach not only ensures that the model output is well-aligned with real-world clinical workflows, but also grounds it in expert-authored, evidence-based content, thereby enabling seamless integration of the models into practical healthcare services.

However, implementing this extension faces three challenges：First, the disease classifications in CPGs are often more detailed than that of the ICD coding system that the models are based on. For example, in the case of hypertensive diseases, CPGs provide detailed classifications that account for variations by clinical severity (such as stage 1 or stage 2) and population-specific factors (including age, or comorbidities, or risk profiles) [9]. In contrast, ICD codes for hypertension are limited to broad categories defined the affected organs or underlying pathological mechanisms [10]. As a result, the output of medical language models often lack the granularity and contextual richness required to align with specific sections of CPGs, leading to overly broad matches that lack specificity and ultimately reduce the practical value of the results.

Second, CPGs are tailored to specific disease domains and are periodically updated by various authoritative healthcare organizations. As a result, numerous versions of guidelines coexist and continue to evolve over time. For a single condition such as hypertension, professional organizations such as the WHO, AAFP, AHA, ASH, ESC, and ESH periodically publish their own treatment guidelines based on the latest clinical research findings [7-9, 11-13]. These guidelines span a wide range of medical specialties, including paediatrics, gynaecology, and cardiology, and are intended for diverse healthcare settings such as primary care facilities and aged care institutions. This dynamic nature of CPGs makes it impractical to include their content into training datasets for fine-tuning language models. Frequent updates and version variations not only risk the generated results to lag behind the latest research findings, but also impose substantial costs for maintaining model relevance through periodic retraining [14].

Third, LLMs are prone to hallucination, generating content that may contain factual inaccuracies, fabrications, or logical inconsistencies [15]. Relying on unverified output from medical language models for disease diagnosis assistance can result in misinformation, misdiagnosis, or unsafe treatment recommendations, posing significant risks to patient safety. The most widely adopted technique to address the hallucination problem is using Retrieval-Augmented Generation (RAG), which dynamically retrieves relevant external medical documents during the generation process to provide supplementary information for content generation [16]. However, even with RAG, the model's output still heavily relies on the parametric knowledge implicitly encoded within the model and therefore cannot guarantee complete consistency with the original content of the retrieved medical documents.

In this paper, we propose a framework called **GARMLE-G** (**G**eneration-**A**ugmented **R**etrieval for **M**edical **L**anguage model **E**xtension – clinical practice **G**uideline) to extend existing medical language models for diagnosis. GARMLE-G innovatively applies Generation-Augmented Retrieval (GAR) mechanism to integrate the medical language model's clinical note comprehension and diagnosis prediction capabilities with the evidence-based, expert-curated knowledge contained in CPGs. By extending existing medical language models this way, GARMLE-G enables the precise delivery of clinical critical information to clinicians, including disease-specific diagnostic criteria, evaluation standards, treatment recommendations, and medication guidelines, thereby providing robust diagnostic support across diverse clinical settings.

This framework effectively addresses the three challenges outlined above. First, it integrates the output of the medical language model with the original EHR input to capture richer information beyond the model output. The enriched data is then encoded into feature embeddings, and by comparing the cosine similarity between feature embeddings, it enables precise matching between the EHR data and the corresponding clinical knowledge within CPGs (Challenge 1). Second, by leveraging the GAR mechanism to dynamically connect to external CPGs, GARMLE-G eliminates the need for retraining or fine-tuning the medical language model as guidelines evolve (Challenge 2). Third, since GAR mechanism retrieves the original CPG contents unmodified as the final output, this framework inherently avoids hallucination by the language model (Challenge 3).

The major contributions of this paper are as follows,

- It proposes GARMLE-G, a GAR-based framework that extends diagnostic medical language models with CPG documents. This framework innovatively incorporates CPGs as external knowledge for medical language models to better address real-world clinical scenarios, while enhancing the credibility of model outputs by mitigating hallucination risks.

- A prototype system specializing in hypertension diagnosis is developed using this framework, which generates CPG-originated diagnostic and treatment recommendations, customized to the specific medical setting and the patient's demographic profile.

- The framework was evaluated through the prototype system. The results indicate that GARMLE-G achieves superior retrieval accuracy, semantic relevance, and guideline adherence compared to conventional RAG-based solutions, while maintaining a lightweight architecture suitable for practical deployment in real-world healthcare environments.

## II. RELATED WORKS

### A. LLMs for Diagnosis

Automated disease diagnosis stands out as a key task of the LLMs in healthcare given its critical role in clinical practice, which enhances diagnostic accuracy, supports physicians in clinical decision-making, and improves the efficiency of healthcare professionals [17, 18]. Most of existing clinical studies apply LLMs by prompt engineering due to its minimal requirement of data and preparation [19, 20]. However, the performance of such method greatly relies on that of the model, where the accuracy of diagnosis is often sub-optimal and unstable. To address this issue, medical language models such as BioBERT, ClinicalBERT, PubMedBERT, Med-PaLM 2 and Meditron [1, 21-24] have been developed, which are fine-tuned from general-purpose base models using domain-specific medical data. Specifically, in [1], it introduced ClinicalBERT, one of the earliest LLMs fine-tuned on clinical notes from the MIMIC-III EHR dataset, demonstrating strong performance across multiple clinical NLP tasks. In [24], Meditron was proposed as a suite of two open-source medical LLMs with different parameter scales, pretrained on a curated corpus comprising PubMed articles and CPGs. The emerge of medical language models marks a significant advancement in applying LLMs to the healthcare domain, but significant limitations remain in their usability and clinical applicability. One major concern is that pretrained models often lack up-to-date medical knowledge, which is critical in rapidly evolving clinical domains [25, 26]. Furthermore, current medical LLMs generally suffer from limited interpretability and hallucinations, posing challenges for clinical adoption where evidence and traceability are essential [27]. Notably, there is a lack of research explicitly integrating CPGs into the training or inference processes of LLMs [28]. This absence significantly hinders alignment with real-world diagnostic workflows, resulting in limited compatibility with evidence-based medical practices.

### B. Optimizing Information Retrieval using External Knowledge

Recent research has explored approaches for LLMs to retrieve up-to-date medical information by integrating external knowledge sources [25, 27, 29-33]. Retrieval-augmented generation (RAG) frameworks are among the most adopted techniques, which retrieves external medical documents relevant to a given query and incorporates this information as additional input to guide LLMs, enabling the generation of up-to-date and accurate output [29, 30]. In addition, a variant of RAG enhances the retrieval process by querying external medical knowledge graphs to acquire domain-specific medical knowledge [25, 31]. For example, in [25], it proposed a method that queries the medical knowledge graph to align medical entities extracted from clinical notes. However, from the perspective of medical diagnostic applications, both frameworks have notable limitations: although external medical knowledge has been incorporated, the output still relies on the generation of LLMs. Such a design hinders improvements in the credibility of model output and fails to effectively address the hallucination problem.

In contrast to RAG, Generation-Augmented Retrieval (GAR) emerge as a technique where LLMs are used not to generate final outputs, but to generate or enhance queries for retrieving external knowledge. The final output comes directly from the retrieved documents rather than the model's generative output [32]. The advantage of GAR lies in the fact that the final output depends on the quality of the retrieved external documents, rather than solely on the generation of the LLM itself. As a result, when credible external documents are available, GAR can produce credible and hallucination-free output. The value of GAR in the medical domain is increasingly being recognized. Several recent studies have adopted this technique as the foundation for their research [27, 32, 33]. Notably, in [27], it proposed the MedRetriever framework, which leverages medical predictive models to retrieve relevant medical text segments crawled from medical websites. These medical text segments and EHR embedding from the patient are then aggregated and used for target disease prediction and interpretation. MedRetriever has demonstrated several effective techniques for retrieving external medical texts using EHRs and LLM output, providing valuable insights and technical foundations that inform our work.

### C. Medical Document-based Information Extraction

Information extraction from medical documents has been an extensively studied area in the field of natural language processing. A wide range of techniques have been proposed for various extraction tasks, including named entity recognition, relation extraction, temporal event identification, and causal inference from sources like EHRs and radiology reports [34-37]. These methods aim to transform unstructured medical narratives into structured representations, which can then be used to support downstream tasks such as clinical decision support, knowledge base construction, and disease risk prediction [38].

CPGs are widely regarded as first-line references in evidence-based medical practice, and are crucial for clinical decision-making, With the rapid advancement of AI technologies in the healthcare sector in recent years, research on automatic information extraction from CPGs has become increasingly important. However, the research topic of information extraction from CPGs remains underexplored [39]. Recent studies have made some attempts to extract structured knowledge from CPGs [40-43]. In [40], the performance of LLMs extracting causal relations from CPGs has been investigated, where BioBERT achieved the best F1-score. In [41], a hybrid approach that integrates input from both machine learning algorithms and domain experts has been proposed for extracting disease-specific information from CPGs. The result demonstrates superior accuracy and efficiency compared to methods relying solely on expert input. Given the inherent complexity and highly specialized nature of CPGs as well as the current state of natural language processing and LLM technologies, fully automated knowledge extraction from CPGs remains a formidable challenge. A promising future direction may lie in expert-annotated, LLM-assisted extraction approaches that combine human expertise with the capabilities of advanced language models.

## III. METHOD

### A. Problem Formulation

GARMLE-G is designed to bridge the gap between diagnosis results generated by medical language models and the expert knowledge embedded in CPGs. Specifically, it aims to align ICD code-based diagnoses produced by the model with the diagnostic and treatment recommendations presented in CPGs, which may appear in various formats including textual descriptions, clinical diagrams, and tables.

In formal representation, let $E = \{e_1, e_2, ..., e_n\}$ denote a patient's EHR record, where each $e_i$ can be a structured data entry, such as a lab test result or an unstructured clinical note. A medical language model $M: E \rightarrow D$ maps $E$ to a set of predicted disease entities $D = \{d_1, d_2, ..., d_k\}$, where each $d_j$ is represented as an ICD code or a textual disease label. Let $G = \{g_1, g_2, ..., g_N\}$ denote the collection of CPG documents. Each document $g_i$ can be segmented into a set of knowledge snippets $S_i = \{s_{i1}, s_{i2}, ..., s_{iM_i}\}$, where $M_i$ is the number of snippets in document $g_i$. The union of all snippets across all CPGs forms the overall snippet pool:

$$S_{all} = \bigcup_{i=1}^{N} S_i = \{s_{ij} | 1 \leq i \leq N, 1 \leq j \leq M_i\} \quad (1)$$

Diagnosis outcome $D$ may lack granularity, contextual depth and adherence to evolving CPG standards. To resolve this, an extended function is $R$ defined as:

$$R(E, D, G) \rightarrow S = \{s_1, s_2, ..., s_m\}, S \subseteq S_{all} \quad (2)$$

where each $s_k \in S$ is a semantically matched snippet retrieved from $S_{all}$ extracted from $G$. Each $s_k$ contains knowledge such as diagnostic criteria, treatment recommendations, or clinical decision pathways relevant to the patient's case.

The problem is therefore formulated as a semantic retrieval task: Given a pair $(E, D)$, retrieve from $S_{all}$ the most relevant set of CPG knowledge entries $S$ by maximizing the semantic similarity $Sim(f(E, D), f(s_i))$, where $f(\cdot)$ is an embedding function mapping text to a feature space, and $Sim(\cdot)$ is a similarity function measures the relevance of two embeddings. Formally, the objective can be expressed as:

$$S = \{s^*\}, where \ s^* = \underset{s_i \in S_{all}}{arg \ max} \ Sim(f(E, D), f(s_i)) \quad (3)$$

In practice, rather than retrieving only the best one match, the top-k most relevant snippets are often retrieved to provide richer retrieval support. Thus, the retrieval function can be further extended as:

$$S = Top_k(\{s_i \in S_{all} | Sim(f(E, D), f(s_i))\}) \quad (4)$$

### B. System Overview

Fig. 1 presents an overview of the GARMLE-G system architecture. To extend existing medical language models with structured CPG knowledge, GARMLE-G operates through three core modules:

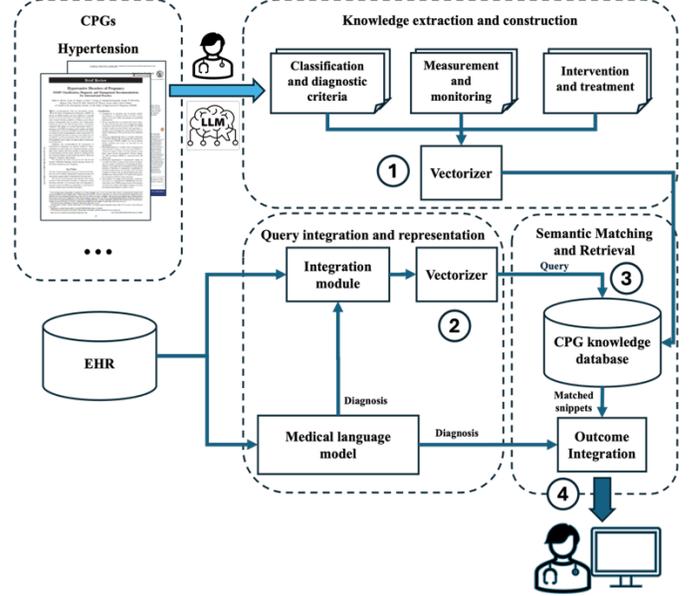

Fig. 1 Overview of GARMLE-G

- **Query Integration and Representation Module** integrates the diagnostic output of the medical language model with the patient's original EHR data to construct an enriched query. The combined input is then transformed into a feature embedding for semantic retrieval.
- **CPG Knowledge Extraction and Construction Module** is responsible for extracting clinically meaningful segments from CPG documents, encoding the segments into feature embeddings, and assemble structured, retrieval-ready CPG snippets.
- **Semantic Matching and Retrieval Module** stores CPG knowledge in the form of structured document snippets and performs similarity-based retrieval. Given a query embedding, it retrieves the most relevant CPG snippets and assembles them with the original model output to generate the final system response.

During actual deployment, the system operates as follows. Step 1, the CPGs are processed by the Knowledge Extraction and Construction Module, where they are transformed into CPG knowledge snippets and stored in the CPG knowledge database. Step 2, when a new diagnostic request is initiated, the Query Integration and Representation Module generates a feature embedding that integrates the medical language model output, the patient's preliminary admission data and historical EHR records, and then sends a query containing this feature embedding to the Semantic Matching and Retrieval Module. Step 3, the Semantic Matching and Retrieval Module processes the query to retrieve the most relevant CPG knowledge snippets. Step 4, the medical language model output is integrated with the retrieved CPG snippets to generate the final system output, which is then presented to clinicians for subsequent medical activities.

### C. CPG Knowledge Extraction and Construction Module

A vast amount of medical knowledge exists across different versions of CPG documents spanning various disease domains. In the context of diagnostic support and downstream clinical tasks, some of this knowledge is particularly valuable, while other portions may be redundant, irrelevant, or unnecessary for extraction. Efficiently extracting the most useful information while minimizing the inclusion of noisy or irrelevant content is critical for maintaining retrieval efficiency and ensuring the quality of the final system outputs. In this section, we describe the CPG knowledge snippet structure designed for knowledge extraction, and the approach used for efficiently extracting and constructing CPG knowledge snippets.

| ID | 1 | 2 | 3 |
|---|---|---|---|
| Text segment embedding | [0.916, -0.743, ..., 0.125] | [0.182,-0.431, ..., 0.025] | [-0.210,0.352, ..., -0.084] |
| Metadata of the snippet | Hypertension: 2023 ESH Guidelines for the management of arterial hypertension:2023:European Society of Hypertension | Hypertension: 2023 ESH Guidelines for the management of arterial hypertension:2023:European Society of Hypertension | Hypertension: 2023 ESH Guidelines for the management of arterial hypertension:2023:European Society of Hypertension |
| Structured text segment | DEFINITION OF HYPERTENSION AND BP CLASSIFICATION:Classification of hypertension:Table 1. Classification of office BP and definitions of hypertension grades | BP MEASUREMENT AND MONITORING:Office BP measurements:Office BP is recommended for diagnosis of hypertension ... | OFFICE BP TARGETS FOR TREATMENT :Patients 18 to 64 years old :The primary goal of treatment is to lower BP to <140/80mmHg. |
| Other relevant values to the snippet | <URL>, ... | CoR: I, LoE: A, ... | CoR: I, LoE: A, ... |

Fig. 2 Snippet structure and examples

Fig. 2 illustrates the data structure of a snippet in CARMLE-G, along with three representative snippet examples. Each snippet consists of five components: the Snippet ID, the Feature Embedding, the Metadata field, the Structured Text field, and Other Related Parameter field. Among them, the Feature Embedding is a high-dimensional dense vector generated by vectorizing the structured text field using a vectorizer. The vectorizer could be a text encoder, a tokenizer-encoder combination, or a lightweight text embedding model. The Metadata field stores information about the source of the snippet, including the original CPG title, CPG version, and publishing organization. The Structured Text field contains textual segments from the CPG documents, with additional annotations such as the corresponding section and sub-section titles, image caption, and table caption. The Other Related Parameters field includes supplementary information necessary for utilizing the snippet but not directly part of the original text fields, such as URLs for tables and images, or the confidence level associated with specific recommendations.

Formally, a CPG document $g_i$ contains a set of knowledge snippets $S_i = \{s_{i1}, s_{i2}, ..., s_{iM_i}\}$, where each snippet $s_{ij}$ is defined as a 5-tuple:

$$s_{ij} = (ID, v_{ij}, M_{ij}, T_{ij}, R_{ij})$$

Where $v_{ij}$ corresponds to embedding for extracted text segment in the snippet; $M_{ij}$ corresponds to metadata; $T_{ij}$ corresponds to the text segment; $R_{ij}$ corresponds to other relevant values. The objective of the CPG knowledge extraction function $\mathcal{E}$ is to select a clinically relevant subset of snippets $S'_i \subseteq S_i$, defined as:

$$S'_i = \{s_{ij} | s_{ij} \in S_i, Relevance(M_{ij}, T_{ij}) \geq \tau\} \quad (5)$$

Where $Relevance(\cdot)$ domain-specific scoring function that quantifies the clinical usefulness of the content of the snippet,

and $\tau$ is the threshold above which a snippet is considered relevant.

Based on discussions with clinicians, the most critical CPG knowledge for diagnostic and downstream clinical tasks falls into three main categories: **Classification and diagnostic criteria**, which define disease entities and guide patient status identification; **Measurement and monitoring**, which standardize physiological assessments and follow-up management; **Intervention and treatment**, which support therapeutic decision-making including drug selection and treatment planning. This information is presented either in the form of text segments, tables, or workflow diagrams, with tables and workflow diagrams often being more efficient for clinical use. Taking the hypertension guideline [13] as an example, the cases shown in Figure 2 illustrate typical snippets representing these three categories from the CPG document. Notably, this guideline adopts the Class of Recommendation (CoR) and Level of Evidence (LoE) grading standards, with the corresponding values stored in the Other Related Parameters field of corresponding snippets.

This module leverages the capabilities of LLMs to assist in the efficient extraction of knowledge and the construction of structured snippets from a large amount of CPG documents. The challenge of ensuring high-quality knowledge extraction and construction lies in generating accurate prompts. To evaluate the performance of different extraction strategies, two approaches are adopted: a hybrid approach that combines expert annotation with LLM-assisted extraction, and a fully automated approach that relies solely on LLMs.

In the hybrid approach, expert knowledge is integrated with the text processing capabilities of LLMs to perform knowledge extraction. First, domain-specific medical professionals—such as general practitioners and specialists familiar with specific disease treatment pathways—are invited to annotate relevant content within the CPG documents. Specifically, experts annotate three types of CPG knowledge: **sub-title of relevant textual segments, number of relevant figures and tables, and the corresponding chapter and section titles where these elements appear**. Second, a prompt template is prepared, which is a framework that contains detailed knowledge extraction requirements. Third, the expert annotations are collected and incorporated into the prompt template. Fourth, the snippet structure is defined and incorporated to finalise the prompt. Finally, the prompt is executed by the LLM, with a set of structured CPG knowledge snippets as output. Table 1 shows a high-level description of the prompt template.

In comparison, the fully automated extraction approach eliminates the need for expert annotation and relies entirely on LLMs for knowledge extraction, with the aim of reducing development costs and minimizing the time burden on medical professionals. Since there are no expert annotations indicating which content should be extracted, the prompt explicitly instructs the LLM to extract only the three main categories of knowledge typically required from a CPG document, as defined earlier. However, due to the substantial variation in writing style and clinical focus across different CPG documents, this fully automated approach proves to be insufficient for extracting structured knowledge accurately. Our experimental results later in this paper show that this approach significantly underperforms the hybrid approach in terms of precision and recall.

Table 1
A HIGH-LEVEL DESCRIPTION OF THE PROMPT TEMPLATE

| Step | Description |
|---|---|
| **Input** | PDF document |
| **Output** | Structured JSON file |
| **1.** Table of Contents Extraction Requirements | Step 1: Title Extraction<br>Step 2: Title Accuracy Validation<br>Step 3: Generate Table of Contents<br>Step 4: Final Check |
| **2.** Snippet format Definition | Extract content from the file using the following structure: <Snippet structure> |
| **3.** Implicit Requirements | <Title Hierarchy><br><Formatting Rules><br><Field Consistency> |
| **4.** Extraction Category | <tagged chapter and section title><br>  <tagged textual segment subtitle>;<br><tagged section and sub-section title><br>  <tagged figure or table number>;<br>… |
| **5.** Final validation | Ensure heading-content consistency;<br>Validate the hierarchical structure and JSON formatting;<br>Remove redundant or irrelevant content; |

The final step of this module involves vectorizing the structured text segments of snippets, i.e., transforming the structured text segments into query embeddings, thereby completing the construction of the CPG knowledge snippets. This step requires the use of the same vectorizer as in Step 2 of Figure 1, to ensure that both query embedding and CPG knowledge embeddings reside in the same semantic space. The vectorizer may take the form of a text-processing encoder or a lightweight LLM. Ideally, it should be fine-tuned for analysing medical text, so that snippet retrieval can be accurately conducted.

### D. Query Integration and Representation Module

After a new diagnostic request is initiated, the base medical language model generates one or more ICD-coded diagnostic outputs based on the patient's EHR data. However, these outputs typically carry limited semantic information, which makes them insufficient to support accurate semantic matching and effective retrieval of CPG knowledge snippets. To address this issue, the module integrates the diagnostic outputs with the patient's EHR textual data—including past medical history—to construct a query enriched with comprehensive clinical context. This query is then transformed into an embedding, which is used for subsequent CPG knowledge matching and retrieval. In this section, we detail the process of integrating diagnostic results with EHR data, as well as the generation of the corresponding embedding.

Formatively, let $D_0$ represent the initial diagnostic output generated by the base medical language model. $E_0$ denotes the content from the EHR record of the current hospital admission. The historical EHR records are denoted as $\{E_{-1}, E_{-2}, \ldots, E_{-n}\}$. Assume that a weighted integration strategy is adopted to combine the current and historical EHR records, capturing temporal relevance while preserving critical background information. The aggregated context $C$ is defined as:

$$C = \omega_0 \cdot E_0 + \sum_{i=1}^{n} \omega_i \cdot E_{-i} \quad (5)$$

where $\omega_0$ is the weight for the current admission record and $\omega_i$ is the weight for the $i$-th most recent historical record. Let the weights follow a normalized time-decay weighting scheme, we have:

$$\omega_i = \frac{e^{-\lambda \cdot \Delta t_i}}{\sum_{j=1}^{n} e^{-\lambda \cdot \Delta t_j}} \quad (6)$$

where $\lambda$ is a tunable hyperparameter controlling the rate of decay and $\Delta t_i$ is the time interval between the $i$-th record and the current time.

The final query $Q$ is then constructed by combining the diagnostic output with the weighted contextual information:

$$Q = [D_0; C] \quad (7)$$

Query $Q$ is subsequently transformed into an embedding using a medical domain vectorizer $f_{embed}(\cdot)$, where the embedding output of the Query Integration and Representation Module

$$q = f_{embed}(Q), q \in \mathbb{R}^d \quad (8)$$

Where $d$ denotes the dimensionality of the embedding space, which is determined by the vectorizer.

In practical implementation, two simplifications were made to the above formalization to better accommodate the content characteristics of EHR data and software-related constraints:

First, considering the input length constraints imposed by the vectorizer, the historical EHR data to be included is simplified. Prior to a definitive diagnosis being made during a hospital admission, the initial patient data available, i.e., the $E_0$ that a hospital collects, typically includes data such as demographic information, chief complaints, past medical history, preliminary laboratory and imaging results, nursing assessment records. Among these, the chief complaint and medical history are often obtained through patient self-report, which introduces a degree of subjectivity. In contrast, historical medical records—particularly physician diagnoses, outpatient visit notes, and discharge summaries—can provide effective and objective supplements to the current admission data, enabling a more comprehensive reconstruction of the patient's health status. Therefore, we simplify each historical EHR record $E_{-i}$ to include only diagnosis, outpatient visit notes and discharge summaries.

Second, to reduce experimental complexity, the prototype employs a fixed-weight integration strategy over a predefined number of historical EHR records. The final query embedding is computed as a weighted sum of the embedding of the current hospital admission and those of the two most recent prior EHR records. The weights used in this integration were empirically determined based on iterative testing to achieve optimal retrieval performance.

E. Query Integration and Representation Module

In this module, the query embedding is sent to the CPG knowledge database. By performing similarity search in the semantic space, the system retrieves the CPG knowledge snippets most relevant to the patient's electronic health record.

These retrieved snippets are then combined with the initial output from the medical language model to generate the final system output. Specifically, this module measures the semantic relevance between the query embedding and CPG knowledge snippet embeddings using cosine similarity. In addition, it adopts a hybrid strategy that combines *top-k* retrieval with a similarity threshold, ensuring that the retrieved results maintain a minimum quality standard while maximizing the number of matched snippets within an acceptable range.

Let the query embedding be $q = f(E, D)$, CPG knowledge snippet embedding be $v_i = f(s_i)$, the snippet embedding collection be defined as $V = \{v_1, v_2, ..., v_n\}$. The semantic similarity to the query $q$ can be calculated as follows:

$$Sim(q, v_i) = \frac{q \cdot v_i}{\|q\| \cdot \|v_i\|} \quad (9)$$

The semantic matching result set $S$ consists of the top-$k$ most relevant snippet that satisfy a minimum similarity threshold τ. Therefore, we denote the set of matched snippets as follows:

$$S = Top_k\left(\{s_i \in S_{all} | Sim(f(E, D), f(s_i)) \geq \tau\}\right) \quad (10)$$

Lastly, the system fuses the diagnostic output $D$ from the base language model with the semantically matched snippet set $S$ to generate the final system response:

$$Output = \Psi(D, S) \quad (11)$$

Here, $\Psi(\cdot)$ denotes the fusion mechanism, which combines $D$ and $S$ in a clinically meaningful manner to produce an output tailored to practical medical applications. In real-world deployments, $\Psi(\cdot)$ may incorporate additional post-processing strategies to further refine the matched results. For example, the function may remove semantically redundant snippets to enhance conciseness and relevance. Moreover, $\Psi(\cdot)$ can adjust the rank of output based on the context in which the diagnostic request is initiated, such as the requesting department, medical specialty, or intended downstream task, thereby enabling context-aware filtering of $S$ to better align with diverse clinical scenarios.

## IV. EXPERIMENTS AND RESULTS

To validate the feasibility of the GARMLE-G framework, we developed a prototype system for hypertension diagnosis and treatment recommendation. The prototype includes a Bio_ClinicalBERT-based diagnosis model to support CPG-driven diagnostic and treatment recommendations for hypertension-related EHR records. The prototype was evaluated on a publicly available EHR dataset from multiple perspectives and compared against state-of-the-art RAG-based enhancement solutions. The experimental setup and evaluation metrics are detailed below.

### A. The Prototype System

The prototype system strictly adheres to the design of the GARMLE-G framework, as illustrated in Figure 1. Several off-the-shelf models and tools are selected for prototype development. Specifically, a fine-tuned Bio-ClinicalBERT model was employed as the base medical language model, capable of generating ICD code-based diagnoses from clinical notes. The ChromaDB vector database was used as the CPG knowledge database, providing native support for cosine similarity-based semantic search and retrieval. The BGE-M3 embedding model served as the vectorizer for both query and document encoding, chosen for its multilingual capability and support for long documents of up to 8K tokens. The entire prototype system was developed and deployed locally for evaluation purposes.

### B. Datasets and Preprocessing

Publicly available EHR dataset and CPG documents are used for the experiment.

The MIMIC VI dataset was used as the source of patient EHR records. It is a large, publicly available EHR dataset that contains de-identified clinical data from over 300,000 patients admitted to the Beth Israel Deaconess Medical Center between 2008 and 2019 [44]. Out of the dataset, hospital records for 13,393 patients were extracted to simulate real-world patient cohorts. Each patient may be associated with 0~5 diagnosis, including hypertension (7,332 cases), hyperlipidemia (4,474 cases), coronary artery disease (3,999 cases), atrial fibrillation (3,592 cases), and others (1,928 cases).

For hypertension-related CPG knowledge extraction, 12 representative hypertension CPGs published between 2016 and 2024 by leading cardiovascular and hypertension research organizations in the United States, Europe, Australia, and the World Health Organization are selected [7, 8, 11-13, 45-50]. Notably, the collection includes two versions of the 2020 guidelines issued by the International Society of Hypertension (ISH), which contain identical content but differ in formatting. These were included to compare the quality and consistency of CPG knowledge extraction across different document structures.

As previously mentioned, this study leverages the capabilities of LLMs to extract CPG knowledge snippets from the selected guideline documents. Two distinct extraction approaches were applied respectively to test their effectiveness and accuracies: a hybrid approach that combines expert annotation with LLM-assisted extraction, and a fully automated approach that relies solely on the language model without human intervention. Both approaches utilized the same prompt template; however, the hybrid approach incorporated human-annotated field tags to guide the extraction process, while the fully automated method only specified the required knowledge categories.

**Table 2** presents a comparison of the two extraction approaches in terms of **precision**, **hit rate**, and **coverage rate**. Specifically, **precision** is defined as the proportion of correctly extracted snippets from the fully automated approach, using the hybrid method as the reference standard (A snippet is considered correct if: I. Its content matches the original CPG text, and II. It aligns with the intended knowledge category）；
Hit rate is defined as the number of auto-extracted snippets that match the human-annotated snippets, divided by the total number of human-annotated snippets (A match is determined when the AI-extracted snippet is either identical to or a sub-span of a human-annotated snippet); Coverage is defined as the total number of snippets extracted by the fully automated approach divided by the total number of human-annotated snippets. As shown in the table, the performance of the fully automated approach is significantly less stable compared to the hybrid method. We observed substantial variation in performance across different CPG documents when using the same prompt, largely due to differences in document formatting and writing styles. Notably, for the two structurally different versions of the same guideline (ISH 2020), the automated extraction approach produced markedly different results. In addition, we found that for several documents, the granularity of automatically extracted snippets was smaller than that of the human-annotated snippets, occasionally leading to a coverage rate exceeding 100%. Furthermore, neither method was able to effectively extract knowledge embedded in tables or figures. Given the critical importance of tabular and visual information in CPGs for downstream clinical tasks, we manually extracted relevant content from tables and figures and incorporated it into the CPG knowledge base.

In total, this experiment utilized 351 knowledge snippets, consisting of text-based snippets extracted using the hybrid method and manually extracted snippets from tables and figures.

Table 2
PERFORMANCE COMPARISON BETWEEN TWO CPG KNOWLEDGE EXTRACTION APPROACHES

| CPG_name | Precision(%) | Hit Rate(%) | Coverage(%) |
|---|---|---|---|
| ISH 2020 v1 | 93.75 | 68.18 | 72.73 |
| ISH 2020 v2 | 89.83 | 81.54 | 90.77 |
| NICE 2023 | 93.1 | 71.05 | 76.32 |
| AAP 2017 | 92.86 | 56.52 | 60.87 |
| ESC 2024 | 90 | 78.26 | 86.96 |
| ESH 2023 | 78.57 | 64.71 | 82.35 |
| ESH 2024 | 93.75 | 93.75 | 100 |
| Heart Fundation 2016 | 96.43 | 90 | 93.33 |
| AAFP 2022 | 100 | 69.23 | 69.23 |
| ISSHP 2018 | 57.78 | 96.3 | 166.67 |
| VADoD 2020 | 84.85 | 58.33 | 68.75 |
| WHO 2020 | 71.79 | 96.55 | 134.48 |

### C. Compared Solutions

This experiment compares the GARMLE-G prototype system with current mainstream approaches that incorporate external knowledge into LLMs via RAG techniques. Given the limited number of LLM platforms that natively support RAG, two representative comparison solutions were selected, which are **ChatGPT-4o + native RAG + PDF and deepseek-7b + AnythingLLM + PDF.** Specifically, ChatGPT-4o provides built-in RAG capabilities that support direct document upload and external text library integration. In contrast, the smaller-scale deepseek-7b model does not natively support RAG functionality. To enable RAG for this model, we integrated an open-source local knowledge base system, AnythingLLM, which facilitates document ingestion and retrieval. For both comparison systems, the CPG documents were uploaded into the corresponding RAG modules, while the EHR data were provided as part of the prompt input. Each model then generated diagnosis and treatment recommendations based on the retrieved CPG content and the EHR context.

### D. Evaluation Metrics

To comprehensively evaluate the performance of GARMLE-G in terms of CPG knowledge retrieval, several standard information retrieval metrics were employed:

- **Precision@K** measures the proportion of correct knowledge snippets retrieved within the top-$K$ results. It is defined as

$$Precision@ = \frac{Number\ of\ correct\ snippets\ in\ top\ K}{K} \quad (12)$$

- **Hits@K** is a relaxed binary metric that measures whether at least one relevant snippet appears in the top-$K$ retrieved results for a given query. It is calculated as:

$$Hits@K = \frac{1}{N}\sum_{i=1}^{N} I_{query_i}(correct\ retrieval\ in\ top\ K) \quad (13)$$

where function I(·) is an indicator function.

- **Mean Reciprocal Rank (MRR)** evaluates the ranking quality by measuring how early the first relevant snippet appears in the candidate list for each query. A higher MRR indicates that the system tends to rank relevant knowledge snippets closer to the top. It is calculated as:

$$MRR = \frac{1}{N}\sum_{i=1}^{N}\frac{1}{Rank_i} \quad (14)$$

where $Rank_i$ denotes the position of the first relevant snippet for the $i$-th query.

Here, how to define **correct retrieval** represents one of the major challenges for evaluation. First, the relevant CPG knowledge varies across different EHR cases, and thus each case requires its own dedicated ground truth set. Second, the outputs generated by LLM+RAG-based comparison systems may not strictly follow the original CPG text, and the retrieved paragraphs may differ in granularity from the pre-extracted snippets in GARMLE-G.

To address these challenges, the following procedures were adopted:

- For each EHR case, a ground truth set was constructed by combining manual annotation with LLM-assisted extraction from the CPG knowledge database. In a randomly sampled subset of 100 EHR cases, the average number of ground truth snippets per case was 102.
- Two complementary definitions of **correct retrieval** were introduced:
  (1) The retrieved paragraph shares at least one sentence that exactly match the original ground truth snippet.
  (2) The BERTScore between the retrieved paragraph and the corresponding ground truth snippet exceeds a predefined threshold.

BERTScore is a semantic similarity metric designed to evaluate the quality of natural language generation by leveraging contextual embeddings from pre-trained language models such as BERT. Based on empirical analysis conducted on a set of sample pairs, the BERTScore threshold was set to 0.72. That is, a retrieval result is considered correct if its BERTScore exceeds 0.72.

### E. Results

Table 3
INFORMATION RETRIEVAL PERFORMANCE – EXACT SENTENCE OVERLAP

| Model | Prec@1 | Prec@3 | Prec@5 | MRR | hits@3 |
|---|---|---|---|---|---|
| Chatgpt-4o | 0.0119 | 0.0119 | 0.0071 | 0.0139 | 0.0119 |

| Deepseek-7b | 0.0119 | 0.004 | 0.0024 | 0.0131 | 0.0119 |

Table 3 summarizes the retrieval performance of the baseline models under the exact sentence overlap criterion, where retrieved snippets are considered correct if they share at least one sentence that exactly match the original ground truth snippet.

Both ChatGPT-4o and Deepseek-7b based solutions exhibited very limited retrieval accuracy under this strict setting. The Precision@1 for both models was 0.0119, with similarly low values across Precision@3, Precision@5, MRR, and Hits@3. In fact, each system correctly retrieved relevant knowledge paragraphs for only one EHR case, where the top 3 retrieved snippets matched the ground truth. Interestingly, further analysis revealed that the low performance was not primarily due to generation-based variations or paraphrasing that failed to exactly match the ground truth. Instead, the main cause was that both models frequently retrieved content originating from different sections of the CPGs compared to the designated ground truth snippets.

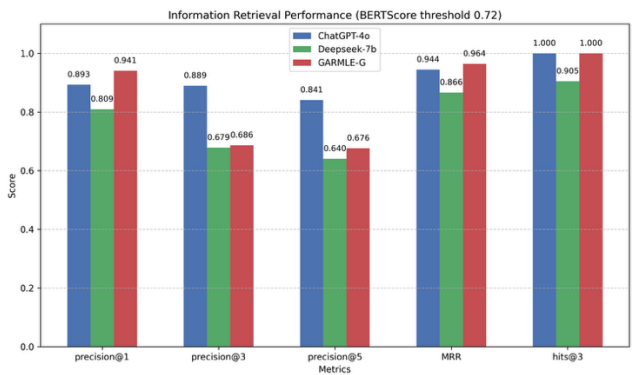

Fig 3 Information Retrieval Performance (BERTScore threshold 0.72)

**Figure 3** presents the retrieval performance of the three models under the BERTScore-based semantic matching criterion (threshold 0.72). Compared to the exact sentence overlap evaluation, both ChatGPT-4o and Deepseek-7b based solutions showed substantial improvement across all metrics. This result suggests that while the knowledge retrieved by these solutions is highly relevant to the target topics, which may or may not be directly derived from the CPG documents, it often does not fully align with the specific content clinicians expect.

On the other hand, our model achieved the highest retrieval performance in Precision@1 of 0.940 and MRR of 0.964, demonstrating strong top-ranked retrieval accuracy and ranking stability. For the metrics of Precision@3 and Precision@5, the ChatGPT-4o based solution achieved the best performance, while CARMLE-G ranked second. In terms of Hits@3, both the ChatGPT-4o based solution and CARMLE-G reached a score of 1.0, indicating that both models consistently retrieved at least one correct CPG knowledge snippet within the top 3 results. Across all metrics, GARMLE-G consistently outperformed the Deepseek-7b-based solution.

These results are particularly encouraging given the significant differences in model scale: while ChatGPT-4o is a massive model containing hundreds of billions of parameters, CARMLE-G achieves comparable retrieval performance using only a lightweight architecture composed of a small BERT-based model, a pair of embedding models totaling 568 million parameters, and a vector database.

### F. Ablation Study

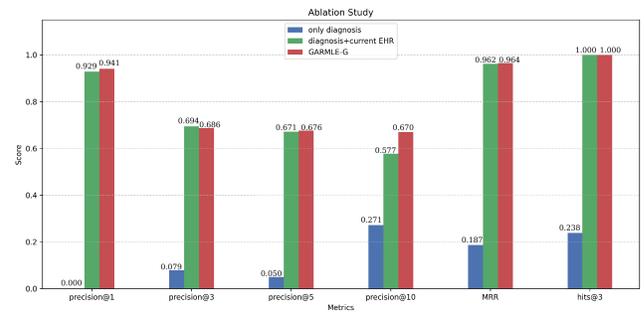

Fig. 4 Ablation study results

An ablation study was conducted to evaluate the contribution of components of GARMLE-G, specifically the integration module that combines the output of the medical language model and the patient EHR records. Three configurations were compared: (1) diagnosis model only, (2) a diagnosis model incorporating EHR information from the current visit, and (3) the full GARMLE-G integrating both current and historical EHR information.

The results demonstrate that incorporating EHR information significantly improves retrieval performance across all evaluation metrics. As shown in Figure 4, the diagnosis model only solution failed to retrieve relevant CPG knowledge in most cases, resulting in a Precision@1 of 0 and an MRR of 0.187. The inclusion of current EHR information significantly improved performance, raising Precision@1 to 0.929 and MRR to 0.962. This highlights the importance of utilizing immediate patient context in guiding accurate CPG knowledge retrieval. Further integrating historical EHR data led to additional performance gains, with the full GARMLE-G solution achieving a Precision@1 of 0.941 and an MRR of 0.964. The performance improvements were also observed in Precision@10, which increased from 0.577 to 0.670 after incorporating historical records. Both the diagnosis+current EHR model and the full GARMLE-G model achieved perfect Hits@3 scores of 1.0, demonstrating stable retrieval of at least one correct CPG knowledge snippet within the top 3 candidates across all test cases. Collectively, these results confirm the effectiveness of both the integration module and the query formulation design in GARMLE-G, which progressively incorporates richer longitudinal patient data to optimize the accuracy of CPG knowledge retrieval

### G. Output example

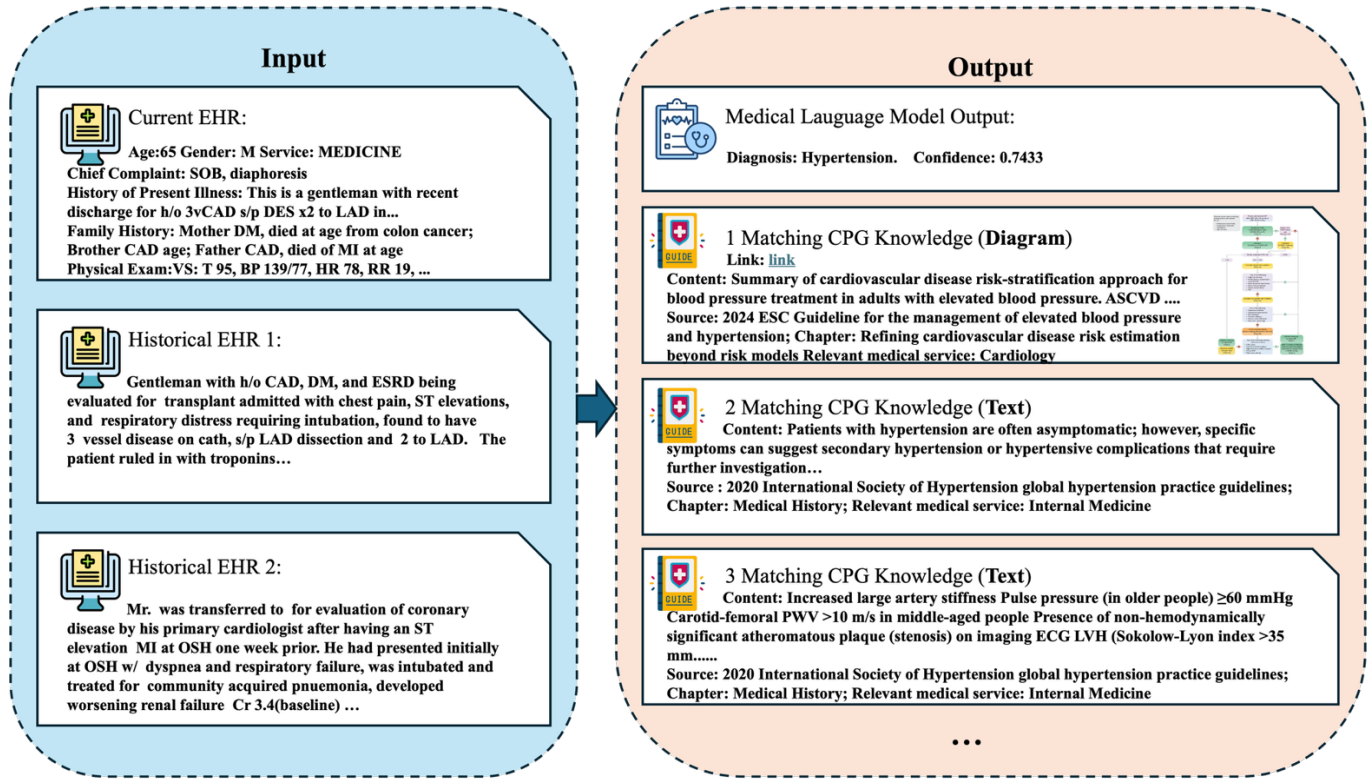

Fig. 5 A demonstrative example of the GARMLE-G prototype

Figure 5 illustrates a representative example to better illustrate the proposed GARMLE-G framework. A 65-year-old male patient presented with shortness of breath (SOB) and diaphoresis. The current EHR record includes clinical notes, family history, and physical examination, while two historical EHR records provide longitudinal information on coronary artery disease (CAD), diabetes mellitus (DM), end-stage renal disease (ESRD), and prior cardiac interventions. Based on the integrated EHR data and medical language model output, the top 3 relevant CPG knowledge retrieved includes a treatment workflow diagram, a guidance on hypertensive symptoms and complications and a treatment recommendation on arterial stiffness and ECG markers of hypertension.

## V. DISCUSSION

In this paper, we present a novel framework that extends medical LLMs by integrating GAR with CPGs. In the proposed framework, the diagnostic output of the language model is enriched with patient-specific EHR data, which is used to retrieve relevant medical knowledge from CPG documents—such as diagnostic criteria, treatment recommendations, and clinical decision pathways. The retrieved guideline-based knowledge is then fused with the model's generation, producing outputs that are both informed by expert consensus and aligned with real-world clinical workflows.

By supplementing medical language model's diagnostic output with authoritative CPG knowledge, the framework enhances the usability of existing models and facilitates seamless integration into contemporary medical workflow processes. Compared with conventional medical language models, this approach offers improved support for downstream clinical tasks, elevating both factual accuracy and adherence to established guidelines. A core innovation of the proposed framework lies in its flexible extension of existing language models: it provides a highly controllable, interpretable, and reliable solution for diagnostic applications. Leveraging the powerful language processing capabilities of LLMs, this framework generates specialized, traceable, and hallucination-free outputs that meet the precision, reliability, and transparency standards demanded in clinical practice.

Importantly, the utility of this extends beyond diagnostic tasks. There is an increasing need to extend medical language models with structured data and domain-specific documents to improve output interpretability and clinical usability [51]. For example, in medication recommendation scenarios, our framework could integrate information from official drug monographs—enabling the model to generate output that include dosage, pharmacological mechanisms, and potential side effects. This capability can obviate the need for manual drug information lookup, thereby enhancing the efficiency and service quality of clinical consultations. Such applications underscore the scalability and practical value of this framework across diverse healthcare domains.

Recent literature underscores that the deployment of medical language models for diagnosis remains in its infancy stages, with real-world adoption constrained by challenges in diagnostic accuracy, clinical ethics, interpretability, and healthcare system integration [52]. Nevertheless, the potential of medical language models in supporting diagnostic reasoning is widely recognized. This study takes an important step toward eliminating the hallucination problem of medical language models, thereby laying a theoretical foundation for their broader clinical application.

Furthermore, while most mainstream LLMs are extremely large in scale, emerging trends indicate a shift toward localization and miniaturization. The release of models such as DeepSeek [53] and ModernBERT [54] exemplifies this shift. In this context, this framework is particularly valuable, providing a lightweight adaptable solution for enhancing small-scale, domain-specific language models deployed in localized medical environments. By applying the GAR-based extension to local models, we can substantially improve the usability and interpretability of generated outputs, enabling seamless integration into hospital workflows and ultimately enhancing the quality of patient care.

## VI. CONCLUSION

This work presents GARMLE-G, a Generation-Augmented Retrieval based framework that extends medical language models with clinical practice guidelines to enhance diagnostic accuracy, and clinical applicability. By incorporating patient-specific EHR data and semantically retrieving guideline-based knowledge, GARMLE-G effectively mitigates hallucination risks and aligns model outputs with real-world medical workflows. Evaluation results demonstrated through a hypertension diagnosis prototype shows that GARMLE-G delivers superior retrieval precision and ranking performance compared to conventional RAG-based solutions, while remaining lightweight and deployable in localized medical environments. In conclusion, the proposed framework offers a practical pathway for integrating medical language models into evidence-based clinical practice.

## ETHICS STATEMENT

This study is conducted using the publicly available and de-identified electronic health record dataset MIMIC-IV. All data used in this study have been de-identified in compliance with the Health Insurance Portability and Accountability Act (HIPAA) standards to ensure patient privacy and confidentiality. Access to the MIMIC-IV database was granted after completion of the required Collaborative Institutional Training Initiative (CITI) "Data or Specimens Only Research" course (Certification ID: 65671049) and successful registration on PhysioNet. As the dataset is fully de-identified and does not involve interaction with human subjects, this research is exempt from institutional review board (IRB) approval. The study complies with ethical principles outlined in the Declaration of Helsinki and follows all applicable data use agreements.

## ACKNOWLEDGMENT

The research reported in this paper is financially supported by the National Natural Science Foundation of China (62276156), the project of Shandong Provincial Natural Science Foundation (ZR2024LZH005), the Taishan Scholar Program of Shandong Province of China (No.tsqnz20240809), and the Excellent Youth Foundation of Shandong Natural Science Foundation (2024HWYQ-055).